\documentclass[10pt,final,journal]{IEEEtran}

\usepackage{amsmath}
\usepackage{amsfonts}
\usepackage{amssymb}
\usepackage{amsthm}
\usepackage{color}
\usepackage{array}
\usepackage{cite}
\usepackage{enumerate}
\usepackage{graphicx}
\usepackage{subfig}
\usepackage{epstopdf}
\usepackage{epsfig}
\usepackage{footmisc}
\usepackage{amsbsy}
\usepackage{bm}
\usepackage{cases}
\usepackage{multirow}
\usepackage{algorithmic}
\usepackage{comment}
\usepackage{times}
\usepackage{paralist}
\usepackage[ruled]{algorithm2e}
\usepackage{setspace}
\usepackage[breakable]{tcolorbox}
\usepackage{kotex}

\newcommand{\bms}{{\bm s}}
\newcommand{\bmz}{{\bm z}}
\newcommand{\bma}{{\bm a}}

\newcommand{\bmdelta}{\bm{\delta}}

\newcommand{\set}[1]{\ensuremath{\mathcal #1}}

\begin{document}
\bstctlcite{IEEEexample:BSTcontrol}

\title{Spiking Neural Networks -- \\ Part I: Detecting Spatial Patterns}

%
%

\author{Hyeryung~Jang, Nicolas~Skatchkovsky, and Osvaldo~Simeone
\thanks{The authors are with the Centre for Telecommunications Research, Department of Engineering, King’s College London, United Kingdom. (e-mail: \{hyeryung.jang, nicolas.skatchkovsky, osvaldo.simeone\}@kcl.ac.uk). 
This work has received funding from the European Research Council (ERC) under the European Union's Horizon 2020 Research and Innovation Programme (Grant Agreement No. 725731).}
}


\maketitle

\begin{abstract}
Spiking Neural Networks (SNNs) are biologically inspired machine learning models that build on dynamic neuronal models processing binary and sparse spiking signals in an event-driven, online, fashion. 
SNNs can be implemented on neuromorphic computing platforms that are emerging as energy-efficient co-processors for learning and inference. 
This is the first of a series of three papers that introduce SNNs to an audience of engineers by focusing on models, algorithms, and applications. 
In this first paper, we first cover neural models used for conventional Artificial Neural Networks (ANNs) and SNNs. 
Then, we review learning algorithms and applications for SNNs that aim at mimicking the functionality of ANNs by detecting or generating spatial patterns in rate-encoded spiking signals. 
We specifically discuss ANN-to-SNN conversion and neural sampling. 
Finally, we validate the capabilities of SNNs for detecting and generating spatial patterns through experiments. 
\end{abstract}

\section{Introduction} \label{sec:introduction}

Artificial Neural Networks (ANNs) have made remarkable progress towards solving difficult tasks through data-driven learning, but most recent successful applications have relied on the availability of massive computing resources \cite{hao2019training}. When considering applications of machine learning for mobile or embedded devices, it is hence both practically and theoretically relevant to explore alternative learning paradigms that may provide more suitable operating points in terms of ``intelligence per joule''. One such alternative framework is given by neuromorphic computing via Spiking Neural Networks (SNNs).

\begin{figure}[t!]
\centering
\includegraphics[height=0.3\columnwidth]{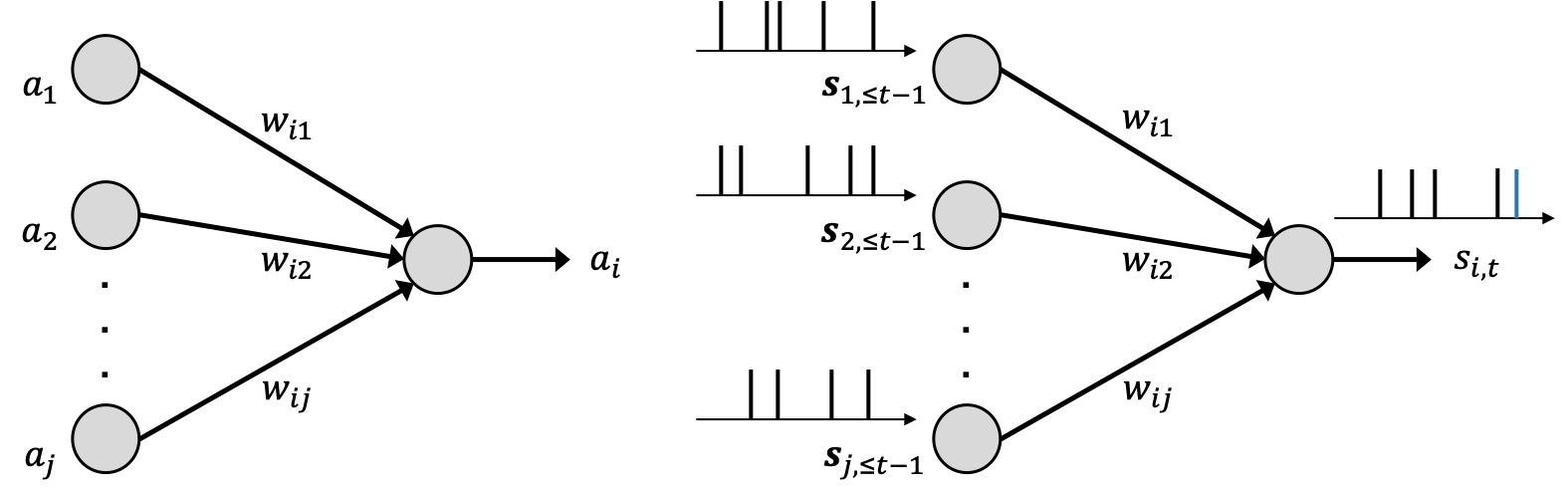}
\vspace{0.2cm}
\caption{Illustration of neural networks: (left) an ANN, where each neuron $i$ processes real numbers $\{a_j\}_{j \in \set{P}_i}$ to output and communicate a real number $a_i$ as a static non-linearity; and (right) an SNN, where each dynamic spiking neuron $i$ outputs and communicates a binary spiking signal $s_{i,t}$ by processing sparse spiking signals $\{\bms_{j,\leq t-1}\}_{j \in \set{P}_i}$ over time $t$.
}
\label{fig:ann_vs_snn}
\end{figure}

SNNs are recurrent networks of dynamic spiking neurons that can process information encoded in the timing of events, or spikes (see Fig. \ref{fig:ann_vs_snn}). SNNs can be implemented on efficient dedicated neuromorphic computing platforms \cite{rajendran2019low}, such as IBM's TrueNorth, Intel's Loihi, and Brainchip's Akida. Using such platforms, recent works have demonstrated practical solutions for specific tasks such as keyword spotting in audio signals \cite{blouw2019benchmarking} and object recognition in videos \cite{chadha2019neuromorphic}, obtaining roughly $5-10 \times$ improvements over conventional ANN deep architecture in terms of energy consumption. Energy consumption in SNNs is essentially proportional to the number of spikes being processed, with each spike requiring as low as a few picojoules \cite{rajendran2019low}.

Due to their potential as energy-efficient co-processors for learning and inference applications on mobile and embedded devices, SNNs may play an important role in the design of communication systems that rely on machine learning. Examples include on-device personalized health assistants trained via federated learning and Internet-of-Things (IoT) monitoring systems for anomaly detection. With the aim of making this topic more easily accessible for researchers in communications, this three-part review paper \cite{snnreviewpt2, snnreviewpt3} will provide an introduction to SNNs with a focus on learning algorithms, applications, and implications on the design of wireless systems. 

Parts I and II describe learning algorithms and applications for conventional tasks, such as memorization, sampling, and classification. Part III covers two use cases of neuromorphic learning for communication systems, namely federated learning for joint training across multiple devices and end-to-end neuromorphic sensing, communication, and remote inference for IoT systems.  In the rest of this section, we outline the content of Part I, and contrast it with that of Part II. 

Neurons in ANNs encode information in \emph{spatial patterns} of real-valued activations. When using a non-negative activation function such as the rectifier in Rectified Linear Units (ReLUs), real-valued activations can be interpreted -- and were originally introduced -- as approximations of the spiking rates of biological neurons. SNNs can mimic more closely the operation of biological neurons by processing directly spiking signals. When information is encoded in the rates of the spiking neurons, an SNN can hence implement the same end-to-end functionality as an ANN but with a potentially lower energy expenditure, a graceful trade-off between performance and operating time, and a  lower latency \cite{davies2018loihi, davies2019spikemark}. In Part I, we will review algorithms and applications based on the detection of spatial patterns via rate encoding. Part II will then concentrate on more advanced solutions that fully leverage the unique capabilities of SNNs to encode information not only in the spiking rates but also in the timing of individual spikes.

The rest of this paper is organized as follows. In Sec.~\ref{sec:spatial-model}, we first cover neural models used in ANNs and SNNs. In Sec.~\ref{sec:spatial-alg}, we introduce learning algorithms for detecting spatial patterns, then describe applications for memorization and sampling tasks in Sec.~\ref{sec:spatial-exp}.





\section{Detecting Spatial Patterns: Models} \label{sec:spatial-model}

\begin{figure}[t!]
\centering
\includegraphics[width=0.8\columnwidth]{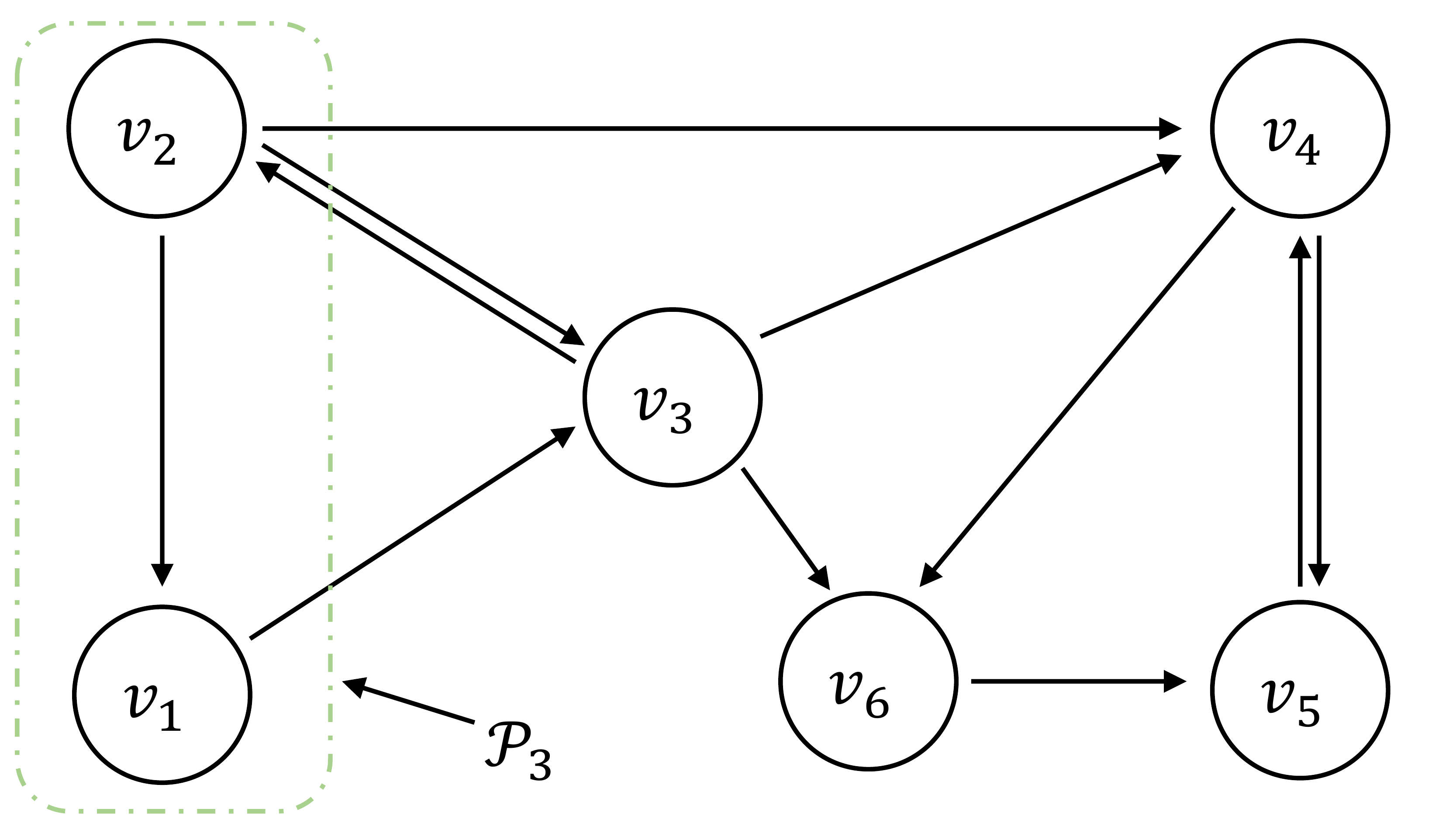}
\vspace{0.2cm}
\caption{Architecture of a SNN with six neurons $\set{V} = \{v_1, v_2, \ldots, v_6\}$. A directed arrow between two neurons represents a synaptic link. This indicates that spikes in the pre-synaptic neuron (origin of the arrow) affect, causally over time, the internal state of the post-synaptic neuron (destination of the arrow). The directed graph may have loops, including self-loops, indicating recurrent behavior.
}
\label{fig:snn_architecture}
\vspace{-0.1cm}
\end{figure}

In this section, we first review the model typically used in standard (non-spiking) ANNs, and then introduce two models that are often used for SNNs, namely the (deterministic) Spike Response Model (SRM) and a simplified version of the SRM with stochastic threshold noise. 

\subsection{ANNs} \label{sec:model-ann}

An ANN is a directed acyclic network of inter-connected neurons, which are typically arranged in a layered architecture. As illustrated in Fig.~\ref{fig:ann_vs_snn}, any neuron $i$ in an ANN is a static unit that outputs a real value $a_i$ in response to real-valued inputs $\bma_{\set{P}_i} = (a_j: j \in \set{P}_i)$ from the set $\set{P}_i$ of ``parent'', or pre-synaptic, neurons. It does so by applying an activation function $f(\cdot)$ as 
\begin{align} \label{eq:ann-output}
    a_i = f(u_i) = f \bigg( \sum_{j \in \set{P}_i} w_{ij} a_j + \gamma_i \bigg),
\end{align}
where $u_i = \sum_{j \in \set{P}_i} w_{ij} a_j + \gamma_i$ is the membrane potential, or pre-activation; $w_{ij}$ is the synaptic weight assigned to the connection between neurons $j$ and $i$; and $\gamma_i$ is a bias. The activation function is often chosen to be a rectifier $f(x) = \max(0,x)$. For an ANN with a layered architecture, the set $\set{P}_i$ of neuron $i$ in layer $l$ consists of the neurons in layer $l-1$.

\subsection{Spike Response Model (SRM) for SNNs} \label{sec:model-SRM}

As illustrated in Fig.~\ref{fig:snn_architecture}, an SNN is a directed, possibly cyclic, network of spiking neurons, where each spiking neuron $i$ is a dynamic system with inputs and outputs given by sequences of spiking signals. Focusing on a discrete-time implementation, as seen in Fig.~\ref{fig:ann_vs_snn}, at each time $t$, any neuron $i$ outputs a binary value $s_{i,t} \in \{0,1\}$, with ``$1$'' denoting the emission of a spike. We collect in vector $\bms_t = (s_{i,t}: i \in \set{V})$ the spikes emitted by the set $\set{V}$ of all neurons at time $t$ and denote by $\bms_{\leq t} = (\bms_1, \ldots, \bms_t)$ the spike sequences up to time $t$. A spike is emitted by neuron $i$ when the {\em membrane potential} $u_{i,t}$ of the neuron at time $t$ crosses a fixed threshold $\vartheta$, after which the membrane potential is reset. Therefore, the binary output $s_{i,t}$ at time $t$ is defined as 
\begin{align} \label{eq:snn-binary}
    s_{i,t} = f(u_{i,t}) = \Theta\big( u_{i,t}-\vartheta \big),
\end{align}
where the activation function $f(\cdot)$ is given by the Heaviside step function $\Theta(\cdot)$ translated by the threshold value $\vartheta$.

To elaborate, we denote as $t_i^{(g)} \in \{0,1,\ldots\}$ with $g=1,2,\ldots$ the sequence of spiking times output by neuron $i$. Furthermore, we introduce the synaptic filter $\alpha_t$, which describes the (impulse) response of a synapse to spikes from a pre-synaptic neuron; and the feedback filter $\beta_t$, which instead models the response to spikes of the neuron itself. We assume strictly causal filters, i.e., $\alpha_t = \beta_t = 0$ for $t \leq 0$. Rather than being a static function of the inputs as in ANNs, the membrane potential of a spiking neuron evolves dynamically in response to the spiking signals of the pre-synaptic neurons and of the neuron itself. The dynamics of the membrane potential $u_{i,t}$ are modeled as
\begin{eqnarray} \label{eq:potential-LIF}
    u_{i,t} &=& \sum_{j \in \set{P}_i} \sum_{g: t_j^{(g)} \leq t} w_{ij} \alpha_{t-t_j^{(g)}} + \sum_{g: t_i^{(g)} \leq t} \beta_{t-t_i^{(g)}} + \gamma_i \cr
    &=& \sum_{j \in \set{P}_i} w_{ij} \big( \alpha_t \ast s_{j,t} \big) + \big( \beta_t \ast s_{i,t}\big) + \gamma_i, 
\end{eqnarray}
where $w_{ij}$ is the synaptic weight from pre-synaptic neuron $j \in \set{P}_i$ to post-synaptic neuron $i$; $\gamma_i$ is a fixed bias that may need to be scaled; and $\ast$ denotes the convolution operator.

An example for the synaptic response is given by the alpha-function $\alpha_t = \exp(-t/\tau_{\text{mem}})-\exp(-t/\tau_{\text{syn}})$, with some positive constants $\tau_{\text{mem}}$ and $\tau_{\text{syn}}$ \cite{gerstner2002spiking}; while a negative feedback filter such as $\beta_t = -\exp(-t/\tau_{\text{ref}})$ is typically selected to model the refractory mechanism upon the emission of a spike, with the constant $\tau_{\text{ref}}$ determining the duration of the refractory period. These responses can be implemented using auto-regressive filters that only require storing two and one, respectively, auxiliary variables \cite{kaiser2020decolle}.

As a special case of the spike response model \eqref{eq:potential-LIF}, accounting only for the first spike by a neuron (and discarding the others) while preventing additional spikes by assuming rectifiers for the filters with sufficiently long refractory period, the membrane potential up to the first spike of neuron $i$ simplifies to \cite{rueckauer2018convtemporal}
\begin{align} \label{eq:potential-ttfs}
    u_{i,t} = \sum_{j \in \set{P}_i: t_j^{(1)} < t} w_{ij} \cdot \big(t-t_j^{(1)}\big) + \gamma_i t.
\end{align}

\subsection{Modeling Refractory Periods with Fixed Duration} \label{sec:model-refractory}

Instead of modeling the refractory effects via a feedback filter as in \eqref{eq:potential-LIF}, an alternative approach is to enforce that, after a spike is emitted by a neuron, no additional spikes are allowed for $\tau_{\text{ref}} > 0$ time steps \cite{buesing11:sampling}. Let the binary variable $z_{i,t} \in \{0,1\}$ equal ``$1$'' if a spike is emitted by neuron $i$ within the time interval $(t-\tau_{\text{ref}}+1, \ldots, t)$, i.e., if $s_{i,t'} = 1$ for some $t' \in \{t-\tau_{\text{ref}}+1, \ldots, t\}$. 
Furthermore, let the categorical variable $\delta_{i,t} \in \{0,1,\ldots,\tau_{\text{ref}}\}$ specify the time remained in the current refractory interval if $z_{i,t} = 1$, i.e., $s_{i,t-(\tau_{\text{ref}}-\delta_{i,t})} = 1$. Accordingly, a spike emitted by the neuron $i$ at time $t$ sets $z_{i,t'} = 1$ for the duration $\tau_{\text{ref}}$ of the refractory mechanism, i.e., for $t' \in \{t, t+1,\ldots,t+\tau_{\text{ref}}-1\}$; and it sets the categorical variable $\delta_{i,t}$ to $\tau_{\text{ref}}$, which counts down to $0$ as $\delta_{i,t'} = \tau_{\text{ref}} - (t'-t)$ for $t' \in \{t, t+1, \ldots, t+\tau_{\text{ref}}-1\}$. We collect in vector $\bmz_t = (z_{i,t}: i \in \set{V})$ and $\bmdelta_t = (\delta_{i,t}: i \in \set{V})$ the binary and categorical vector defined by the activities of all neurons $\set{V}$ at time $t$.

\subsection{SRM with Stochastic Threshold Noise} \label{sec:model-sampling}

In this section, we review a simplified version of the SRM with stochastic threshold noise that can be used for Bayesian Monte Carlo inference, as we will describe in the next section. A more general version of the model will be discussed in Part II. 
We consider the fixed-duration refractory period model introduced above, so that the dynamics of the system are described by the set of variables $(\bmz_t, \bmdelta_t)$. 

As a simplified version of the SRM \eqref{eq:potential-LIF}, the membrane potential for neuron $i$ is given as the linear weighted sum
\begin{align} \label{eq:potential-sampling}
    u_{i,t} = \sum_{j \in \set{P}_i} w_{ij} z_{j,t} + \gamma_i,
\end{align}
where the contribution $z_{j,t}$ from pre-synaptic neuron $j$ approximates the response of the synapse; and $\gamma_i$ is a constant bias. 
Furthermore, instead of relying on a deterministic threshold activation function, the model assumes that the probability of spiking at time $t$ for neuron $i$ is given as
\begin{align} \label{eq:spiking-sampling}
    p(z_{i,t} = 1|\bmz_{\setminus i, t}) = \sigma(u_{i,t}),
\end{align}
where $\bmz_{\setminus i,t}$ collects the binary values $z_{j,t}$ of all other neurons $j \neq i$ at time $t$, and $\sigma(x) = 1/(1+\exp(-x))$ is the sigmoid function.

\section{Detecting Spatial Patterns: Algorithms} \label{sec:spatial-alg}

In this section, we describe learning algorithms that aim at detecting spatial patterns in the input, hence mimicking the operation of ANNs. 

\subsection{ANN-to-SNN Conversion}
\label{sec:ann-to-snn}
When information is encoded in the rate of the spiking signals, an SNN can approximate the behavior of a conventional ANN with non-negative activation functions such as for ReLU by transferring weights from a pre-trained ANN \cite{rueckauer2017convrate, cao2015convcnn, rueckauer2018convtemporal}. This is done by matching the non-negative activations of the neurons in an ANN with the spiking rates of the corresponding spiking neurons in an SNN. Alternatively, one could map the non-negative activation of a converted neurons to the timing of a {\em single spike} emitted by a corresponding spiking neuron. In this case, it is possible to train an auxiliary non-spiking neural network in which the activation functions approximate the spike timing of a spiking neuron, and then transfer the weights for use in an SNN \cite{comsa2020temporal}.

Consider first the case of rate encoding. To start, we train an ANN with ReLU activation function $f(x) = \max(0,x)$. The conversion mechanism introduced in \cite{rueckauer2017convrate} assumes an SRM in which the synaptic responses are given by the exponential filter $\alpha_t = \exp(-t)$ and a feedback filter $\beta_t = -\vartheta \cdot \delta(t-1)$, with $\delta(\cdot)$ being Kronecker delta function, i.e., $\delta(t) = 1$ for $t=0$ and $\delta(t) = 0$ otherwise.  
For any threshold value $\vartheta$, the dynamics of the spiking neurons in \eqref{eq:potential-LIF} with these choices produce spiking sequences $\bms_{\leq t}$ with an average rate that is related to the corresponding average activation $a_i$ in the ANN. Following \cite{rueckauer2017convrate}, the spiking rate $r_{i,t}$ of neuron $i$ at time $t$ is specifically given as 
\begin{align} \label{eq:conv-rate}
    r_{i,t} = a_i - \frac{u_{i,t}}{t \cdot \vartheta}
\end{align}
for a sufficiently large $t$. From \eqref{eq:conv-rate}, the rate coding-based conversion becomes more accurate as the SNN is operated for a longer period of time $t$, yielding a trade-off among accuracy, latency, and energy expenditure through the number of spikes.

A potentially more efficient operation is obtained by encoding information in the first spiking time of the neurons -- an approach known as time-to-first-spike (TTFS) \cite{rueckauer2018convtemporal}. Considering the membrane potential model in \eqref{eq:potential-ttfs}, the timing of the first spike $t_i^{(1)}$ of neuron $i$ is determined when the membrane potential equals to the threshold, i.e., $u_{i,t_i^{(1)}} = \vartheta$, as
\begin{align} \label{eq:ttfs}
    t_i^{(1)} = \frac{1}{\sum_{j \in \set{P}_i} w_{ij} + \gamma_i} \bigg( \vartheta + \sum_{j \in \set{P}_i: t_j^{(1)} < t} w_{ij} t_j^{(1)} \bigg).
\end{align}
By replacing the constant threshold $\vartheta$ by a dynamic threshold that adapts to the input spikes, as proposed in \cite{rueckauer2018convtemporal}, it is possible to ensure that the spike rate $1/t_i^{(1)}$ approximately matches to the corresponding activation $a_i$ of the ANN.

\subsection{Neural Sampling} \label{sec:neural-sampling}

\begin{figure}[t!]
\centering
\includegraphics[height=0.65\columnwidth]{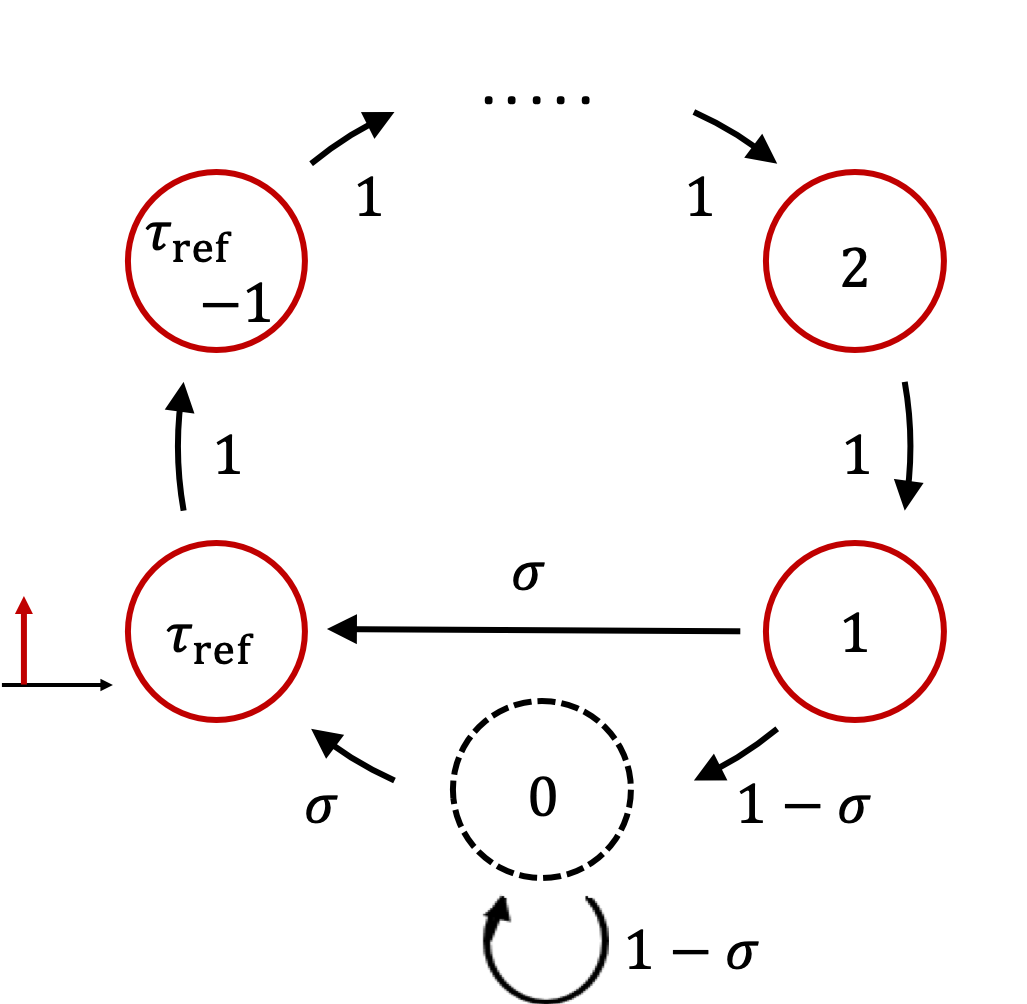}
\vspace{0.3cm}
\caption{Illustration of a schematic of neural sampling (adapted from \cite{buesing11:sampling}). The arrows represent the transition probabilities for the Markov chain associated with the categorical variable $\delta_i$ of a spiking neuron $i$ that has a refractory period of fixed duration $\tau_{\text{ref}}$. The values inside the circles correspond to $\delta_{i,t}$, while the values on top of the arrows represent transition probabilities. The neuron emits a spike with probability $\sigma := \sigma(u_{i,t}-\log \tau_{\text{ref}})$ at time $t$ if it is in the last refractory state $\delta_{i,t-1} = 1$ or in the resting state $\delta_{i,t-1} = 0$ at time $t-1$. We have $z_{i,t} = 1$ for all states $\delta_{i,t} \geq 1$ (solid circle) and $z_{i,t} = 0$ for $\delta_{i,t} = 0$ (dashed circle).
}
\label{fig:neural_sampling}
\vspace{-0.1cm}
\end{figure}

While ANN-to-SNN conversion aims at mimicking the end-to-end operation of an ANN, we now consider an approach that leverages SNNs to tackle a key problem in Bayesian inference, namely sampling from a Boltzmann distribution \cite{buesing11:sampling}. For a given directed graph as described in Sec.~\ref{sec:spatial-model}, consider the Boltzmann distribution 
\begin{align} \label{eq:BM}
    p(\bmz) \propto \exp\bigg( \sum_{i \in \set{V}, j \in \set{P}_i} \frac{1}{2} w_{ij} z_{i,t} z_{j,t} + \sum_{i \in \set{V}} \gamma_i z_{i,t} \bigg),
\end{align}
for some parameters $\{w_{ij}\}$ and $\{\gamma_i\}$, over a binary vector $\bmz=(z_i; i \in \set{V})$, where we have omitted the normalization constant. 
Neural sampling implements a Markov Chain Monte Carlo algorithm that asymptotically produces samples from the distribution \eqref{eq:BM}. As shown in \cite{buesing11:sampling} and reviewed below, this can be done by running the SNN model described in Sec.~\ref{sec:model-sampling}.

Under this model, as illustrated in Fig.~\ref{fig:neural_sampling}, the dynamics of the joint variables $(\bmz_t, \bmdelta_t)$ at time $t$ described by \eqref{eq:potential-sampling}-\eqref{eq:spiking-sampling} can be interpreted as one step of a Markov chain. Refer to $(z_t, \delta_t)$ as the current state of the Markov Chain. If neuron $i$ is in a refractory state at time $t$, i.e., if $\delta_{i,t-1} > 1$ and $z_{i,t-1} = 1$, a deterministic transition takes place to the state $\delta_{i,t} = \delta_{i,t-1}-1$ and $z_{i,t} = 1$. If neuron $i$ is not in refractory at time $t$, i.e., if either $\delta_{i,t-1} = 1$ and $z_{i,t-1}=1$ or $\delta_{i,t-1} = 0$ and $z_{i,t-1} = 0$, the neuron emits a spike with probability $\sigma\big(u_{i,t} - \log \tau_{\text{ref}}\big)$, and a transition to $z_{i,t} = 1$ and $\delta_{i,t} = \tau_{\text{ref}}$ is accordingly carried out. 
In \cite{buesing11:sampling}, it is shown that the resulting Markov chain is aperiodic and irreducible, and that it has a unique invariant distribution $p(\bmz, \bmdelta)$, with the marginal $p(\bmz) = \sum_{\bmdelta} p(\bmz,\bmdelta)$ being the Boltzmann distribution in \eqref{eq:BM}. This ensures that after a sufficiently large time, the probabilistic neural dynamics of the SNN produce samples $\bmz$ from the given Boltzmann distribution $p(\bmz)$ in \eqref{eq:BM}.

\section{Applications} \label{sec:spatial-exp}

\subsection{Image Classification}
Traditional image datasets have often been used in the SNN literature to test the performance of training algorithms \cite{diehl2015fast, hu2018spiking, lee2016training, sengupta2019going, wu2018spatio} by adopting rate encoding. Accordingly, each pixel is encoded into a spiking signal with a spiking probability given by the (normalized) pixel intensities. The spiking signal is generated following an i.i.d. Bernoulli process with the spiking probability -- a process known as Poisson encoding. Alternatively, the pixel intensity can be directly encoded as an analog, constant input \cite{rueckauer2017convrate}.

\begin{figure}
\centering
\includegraphics[scale=0.27]{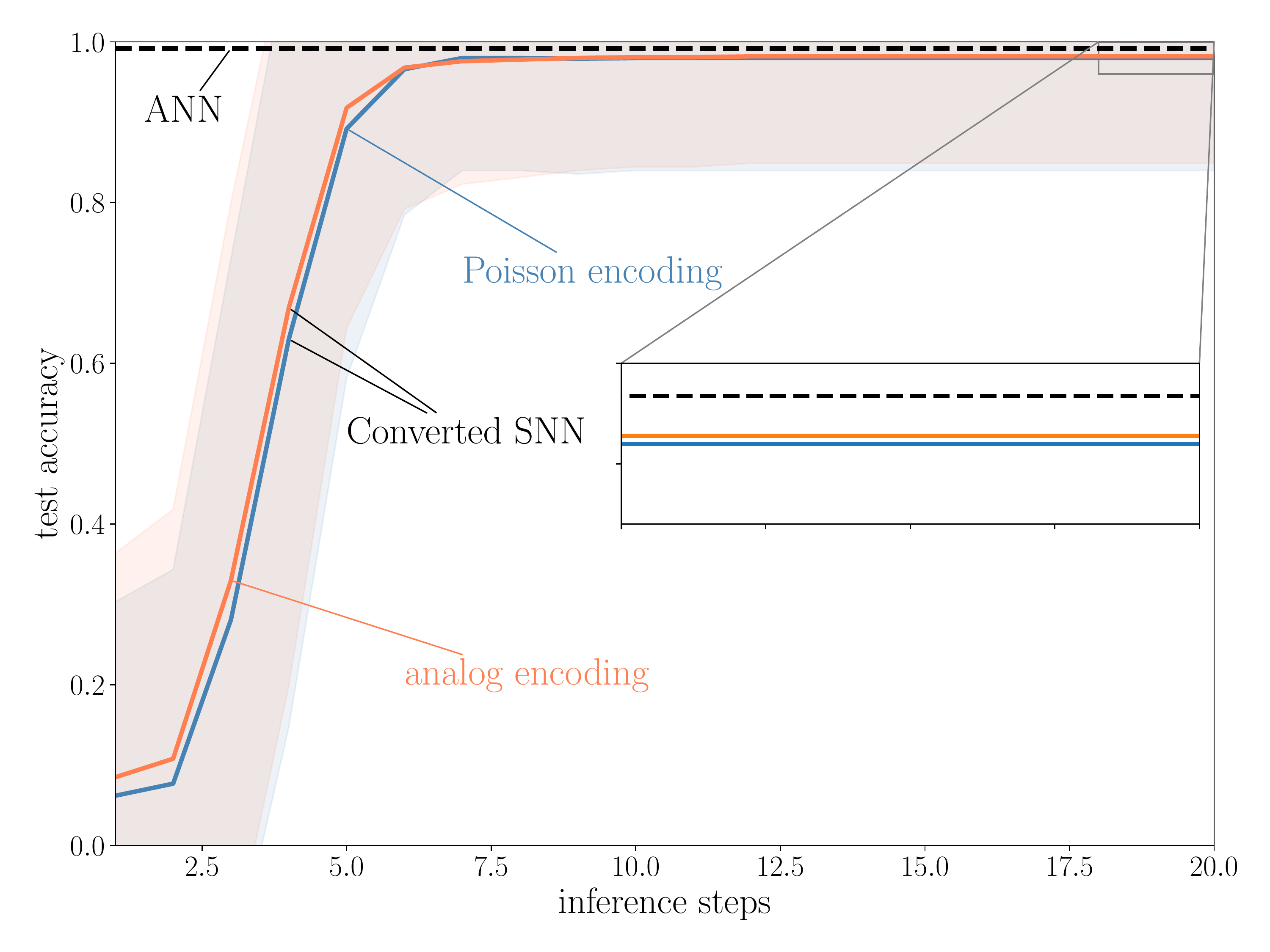}
\caption{Test accuracy as a function of the number of inference time-steps. Shaded areas represent the standard deviation over the test dataset.}
\label{fig:acc_per_timesteps}
\vspace{-0.1cm}
\end{figure}

We evaluate the performance of the rate encoding-based ANN-to-SNN conversion described in Sec.~\ref{sec:ann-to-snn} \cite{rueckauer2017convrate} on the MNIST dataset \cite{lecun-mnisthandwrittendigit-2010}. For this purpose, we used the standard Convolutional Neural Network (CNN) LeNet-5 \cite{lecun1998lenet}, which is trained on the $60,000$ images of the training dataset, and tested on the remaining $10,000$ images. We compare the classification test accuracy of the ANN with the accuracy attained by the converted SNN.

Fig.~\ref{fig:acc_per_timesteps} shows the classification test accuracy attained by the SNN as a function of the inference time-step, that is, of the number of processed input samples. The figure demonstrates that SNNs provide a graceful time-to-accuracy performance, whereby the prediction accuracy improves as the number of processed time samples increases. This is in contrast to ANNs for which a prediction is obtained by processing the entire input signal at once. We also note that the converted SNNs suffer from a minor precision loss as compared to the ANN, and that the network with analog inputs slightly outperform its counterpart with Poisson encoding.

\subsection{Neural Sampling}
\begin{figure}[t!]

\subfloat{%
  \includegraphics[clip,width=\columnwidth]{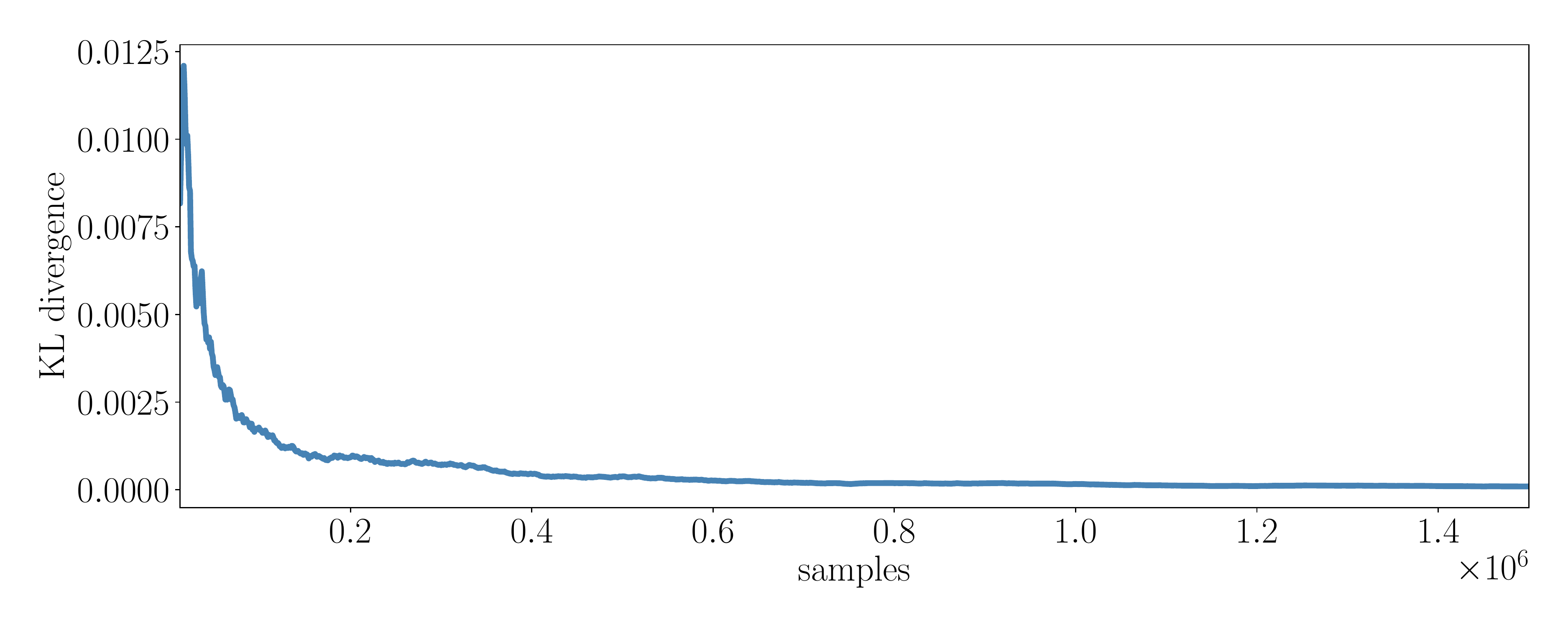}%
}
\vspace{-0.2cm}
\subfloat{%
  \includegraphics[clip,width=\columnwidth]{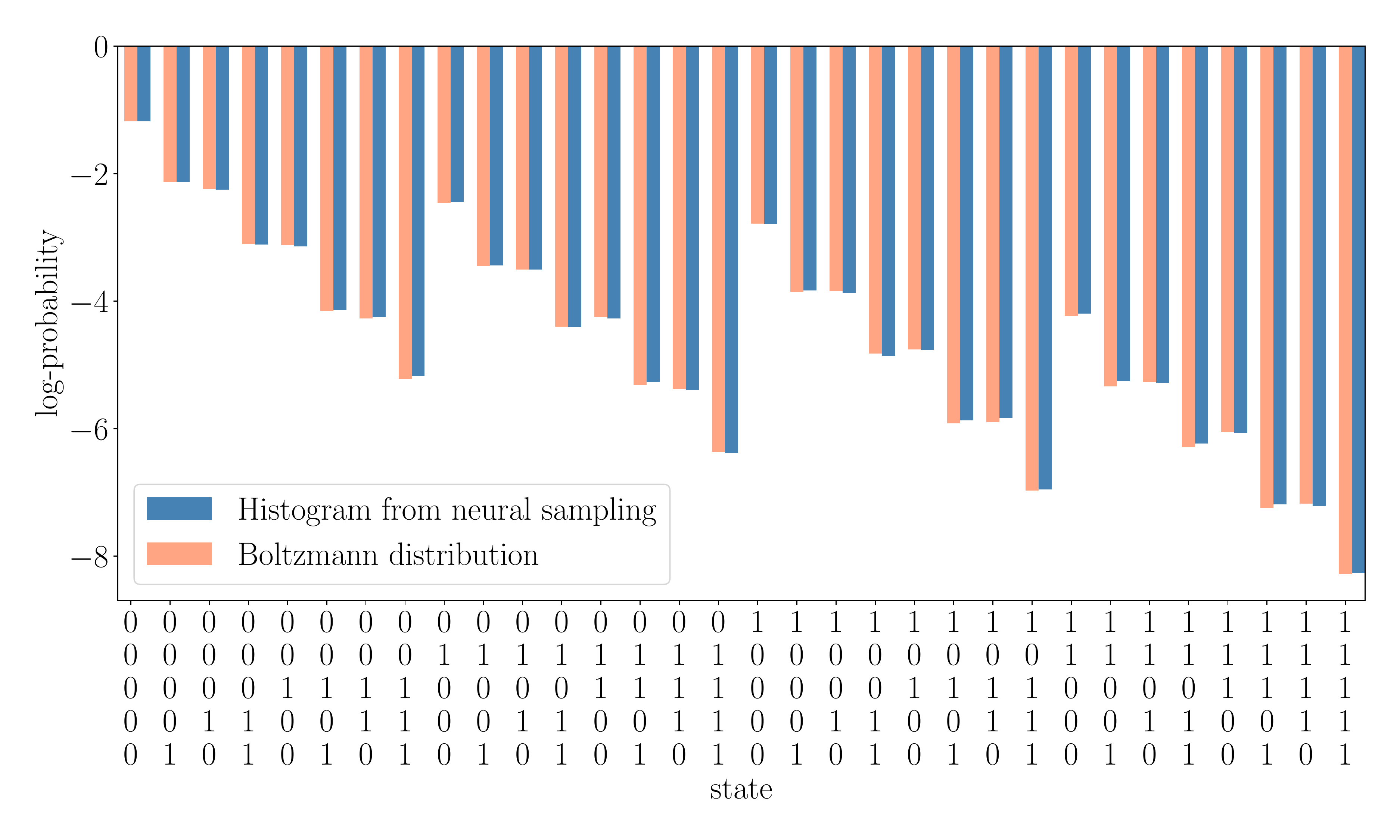}%
}
\caption{Illustration of neural sampling. Top: KL divergence between the histogram of the samples produced by the SNN via neural sampling, and Boltzmann distribution \eqref{eq:BM} as a function of the number of samples. Bottom: Histogram obtained from neural sampling over $1.5 \times 10^6$ time steps and Boltzmann distribution \eqref{eq:BM}. Columns represent the different states taken by the binary variables.}
\label{fig:neural-sampling}
\vspace{-0.1cm}
\end{figure}

In this section, we study the problem of training the GLM-based SNN model described in Sec.~\ref{sec:model-SRM}. 
We evaluate the capability of SNNs to sample from the Boltzmann distribution \eqref{eq:BM}. We consider a network with five fully connected neurons, so that the set of parents for all neurons is given as $\mathcal{P}_i = \mathcal{V} \setminus \{i\} $. The synaptic weights $w_{i,j}$ are symmetric, i.e. $w_{i,j} = w_{j,i}$, and independently sampled from distribution $\mathcal{N}(0, 0.3^2)$, and the biases $\gamma_i$ are independently sampled from distribution $\mathcal{N}(-1.5, 0.5^2)$. After firing, a neuron enters a refractory state for $\tau_\text{ref} = 10$ time-steps. The network is simulated for $T = 1.5 \times 10^6$ time steps, and the state of the network is recorded to evaluate the empirical distribution over the 32 configurations of the $5$ binary variables of interest. 

In Fig~\ref{fig:neural-sampling}, we plot the Kullback-Leibler (KL) divergence between the Boltzmann distribution \eqref{eq:BM} and the histogram of the samples generated so far by the SNN, as a function of the number of samples. The figure shows that, after a sufficient burn-in period, the histogram converges towards a probability distribution that closely matches the Boltzmann distribution \eqref{eq:BM}. 
This is further confirmed by the bottom figure, in which we compare the log-probability of the Boltzmann distribution \eqref{eq:BM} with the logarithm of the histogram of the outputs of the SNN evaluated after $1.5 \times 10^6$ samples. Overall, these results validate neural sampling via SNNs as a viable alternative to classical sampling algorithms such as Markov Chain Monte Carlo (MCMC) methods for Boltzmann distributions. Further refinements, such as relative refractory mechanisms, can be used to improve the performance \cite{buesing11:sampling}.  

\section{Conclusions} \label{sec:conclusion}
In this paper, the first of a three-part review on SNNs, we introduced the Spike Response Model (SRM), along with a probabilistic variant, and reviewed two learning algorithms aimed at detecting or generating spatial patterns with SNNs. The first converts trained ANNs into spiking models. The second, known as neural sampling, tackles the problem of sampling from a Boltzmann distribution using SNNs. Through experiments, we validated the capability of SNNs of closely matching the performance of state-of-the-art algorithms for detection and sampling. In the next part, we concentrate on more advanced solutions  that fully leverage the unique capabilities of SNNs to encode information not only in the  spiking  rates but also in the timing of individual spikes.





\bibliographystyle{IEEEtran}
\bibliography{ref}

\begin{thebibliography}{10}
\providecommand{\url}[1]{#1}
\csname url@samestyle\endcsname
\providecommand{\newblock}{\relax}
\providecommand{\bibinfo}[2]{#2}
\providecommand{\BIBentrySTDinterwordspacing}{\spaceskip=0pt\relax}
\providecommand{\BIBentryALTinterwordstretchfactor}{4}
\providecommand{\BIBentryALTinterwordspacing}{\spaceskip=\fontdimen2\font plus
\BIBentryALTinterwordstretchfactor\fontdimen3\font minus
  \fontdimen4\font\relax}
\providecommand{\BIBforeignlanguage}[2]{{%
\expandafter\ifx\csname l@#1\endcsname\relax
\typeout{** WARNING: IEEEtran.bst: No hyphenation pattern has been}%
\typeout{** loaded for the language `#1'. Using the pattern for}%
\typeout{** the default language instead.}%
\else
\language=\csname l@#1\endcsname
\fi
#2}}
\providecommand{\BIBdecl}{\relax}
\BIBdecl

\bibitem{hao2019training}
K.~Hao, ``Training a single {AI} model can emit as much carbon as five cars in
  their lifetimes,'' \emph{MIT Technology Review}, 2019.

\bibitem{rajendran2019low}
B.~Rajendran, A.~Sebastian, M.~Schmuker, N.~Srinivasa, and E.~Eleftheriou,
  ``Low-power neuromorphic hardware for signal processing applications: A
  review of architectural and system-level design approaches,'' \emph{IEEE
  Signal Processing Magazine}, vol.~36, no.~6, pp. 97--110, 2019.

\bibitem{blouw2019benchmarking}
P.~Blouw, X.~Choo, E.~Hunsberger, and C.~Eliasmith, ``Benchmarking keyword
  spotting efficiency on neuromorphic hardware,'' in \emph{Proc. of Annual
  Neuro-inspired Computational Elements Workshop}.\hskip 1em plus 0.5em minus
  0.4em\relax ACM, 2019.

\bibitem{chadha2019neuromorphic}
A.~Chadha, Y.~Bi, A.~Abbas, and Y.~Andreopoulos, ``Neuromorphic vision sensing
  for cnn-based action recognition,'' in \emph{Proc. of International
  Conference on Acoustics, Speech and Signal Processing}.\hskip 1em plus 0.5em
  minus 0.4em\relax IEEE, 2019.

\bibitem{snnreviewpt2}
N.~Skatchkovsky, H.~Jang, and O.~Simeone, ``Spiking neural networks -- {P}art
  {II}: Detecting spatio-temporal patterns,'' submitted.

\bibitem{snnreviewpt3}
N.~Skatchkovsky, H.~Jang, and O.~Simeone, ``Spiking neural networks -- {P}art
  {III}: Neuromorphic communications,'' submitted.

\bibitem{davies2018loihi}
M.~Davies~et al., ``Loihi: A neuromorphic manycore processor with on-chip
  learning,'' \emph{IEEE Micro}, vol.~38, no.~1, pp. 82--99, 2018.

\bibitem{davies2019spikemark}
M.~Davies, ``Benchmarks for progress in neuromorphic computing,'' \emph{Nature
  Machine Intelligence}, vol.~1, no.~9, pp. 386--388, 2019.

\bibitem{gerstner2002spiking}
W.~Gerstner and W.~M. Kistler, \emph{Spiking neuron models: Single neurons,
  populations, plasticity}.\hskip 1em plus 0.5em minus 0.4em\relax Cambridge
  University Press, 2002.

\bibitem{kaiser2020decolle}
J.~Kaiser, H.~Mostafa, and E.~Neftci, ``Synaptic plasticity dynamics for deep
  continuous local learning ({DECOLLE}),'' \emph{Frontiers in Neuroscience},
  vol.~14, p. 424, 2020.

\bibitem{rueckauer2018convtemporal}
B.~Rueckauer and S.-C. Liu, ``Conversion of analog to spiking neural networks
  using sparse temporal coding,'' in \emph{Proc. of International Symposium on
  Circuits and Systems}, 2018.

\bibitem{buesing11:sampling}
L.~Buesing, J.~Bill, B.~Nessler, and W.~Maass, ``Neural dynamics as sampling: a
  model for stochastic computation in recurrent networks of spiking neurons,''
  \emph{PLoS Computational Biology}, vol.~7, no.~11, p. e1002211, 2011.

\bibitem{rueckauer2017convrate}
B.~Rueckauer, I.-A. Lungu, Y.~Hu, M.~Pfeiffer, and S.-C. Liu, ``Conversion of
  continuous-valued deep networks to efficient event-driven networks for image
  classification,'' \emph{Frontiers in Neuroscience}, vol.~11, p. 682, 2017.

\bibitem{cao2015convcnn}
Y.~Cao, Y.~Chen, and D.~Khosla, ``Spiking deep convolutional neural networks
  for energy-efficient object recognition,'' \emph{International Journal of
  Computer Vision}, vol. 113, no.~1, pp. 54--66, 2015.

\bibitem{comsa2020temporal}
I.~M. Comsa \emph{et~al.}, ``Temporal coding in spiking neural networks with
  alpha synaptic function,'' in \emph{Proc. of International Conference on
  Acoustics, Speech and Signal Processing}.\hskip 1em plus 0.5em minus
  0.4em\relax IEEE, 2020.

\bibitem{diehl2015fast}
P.~U. Diehl \emph{et~al.}, ``Fast-classifying, high-accuracy spiking deep
  networks through weight and threshold balancing,'' in \emph{Proc. of
  International Joint Conference on Neural Networks}.\hskip 1em plus 0.5em
  minus 0.4em\relax IEEE, 2015.

\bibitem{hu2018spiking}
Y.~Hu, H.~Tang, Y.~Wang, and G.~Pan, ``Spiking deep residual network,''
  \emph{arXiv preprint arXiv:1805.01352}, 2018.

\bibitem{lee2016training}
J.~H. Lee, T.~Delbruck, and M.~Pfeiffer, ``Training deep spiking neural
  networks using backpropagation,'' \emph{Frontiers in Neuroscience}, vol.~10,
  p. 508, 2016.

\bibitem{sengupta2019going}
A.~Sengupta, Y.~Ye, R.~Wang, C.~Liu, and K.~Roy, ``Going deeper in spiking
  neural networks: Vgg and residual architectures,'' \emph{Frontiers in
  Neuroscience}, vol.~13, p.~95, 2019.

\bibitem{wu2018spatio}
Y.~Wu, L.~Deng, G.~Li, J.~Zhu, and L.~Shi, ``Spatio-temporal backpropagation
  for training high-performance spiking neural networks,'' \emph{Frontiers in
  Neuroscience}, vol.~12, 2018.

\bibitem{lecun-mnisthandwrittendigit-2010}
\BIBentryALTinterwordspacing
Y.~LeCun and C.~Cortes, ``{MNIST} handwritten digit database,'' 2010. [Online].
  Available: \url{http://yann.lecun.com/exdb/mnist/}
\BIBentrySTDinterwordspacing

\bibitem{lecun1998lenet}
Y.~LeCun \emph{et~al.}, ``Backpropagation applied to handwritten zip code
  recognition,'' \emph{Neural Computation}, vol.~1, no.~4, pp. 541--551, 1989.

\end{thebibliography}


\end{document}